\def\year{2022}\relax
\documentclass[letterpaper]{article} 
\usepackage{aaai22}  
\usepackage{times}  
\usepackage{helvet}  
\usepackage{courier}  
\usepackage[hyphens]{url}  
\usepackage{graphicx} 
\usepackage{natbib}  
\usepackage{caption} 
\DeclareCaptionStyle{ruled}{labelfont=normalfont,labelsep=colon,strut=off} 
\usepackage{multirow}
\usepackage{multicol}
\usepackage{subfigure}
\usepackage{makecell}
\usepackage{algorithm}
\usepackage{amsmath}   

\usepackage{algorithm}
\usepackage{algorithmic}
\usepackage{amssymb}
\usepackage{color}
\usepackage{newfloat}
\usepackage{listings}
\floatstyle{ruled}
\newfloat{listing}{tb}{lst}{}
\floatname{listing}{Listing}
\title{Transcoded Video Restoration by Temporal Spatial Auxiliary Network} 
\author{
    Li Xu\textsuperscript{\rm 1},
    Gang He\textsuperscript{\rm 1,2}\thanks{Corresponding author.},
    Jinjia Zhou\textsuperscript{\rm 3},
    Jie Lei\textsuperscript{\rm 1},
    Weiying Xie\textsuperscript{\rm 1},
    Yunsong Li\textsuperscript{\rm 1},
    Yu-Wing Tai\textsuperscript{\rm 2}
}

\affiliations{
    \textsuperscript{\rm 1}Xidian University, China \ \ 
     \textsuperscript{\rm 2}Kuaishou Technology, China \ \ 
      \textsuperscript{\rm 3}Hosei University, Japan\\

    cherylxu@stu.xidian.edu.cn, \{ghe, wyxie\}@xidian.edu.cn, zhou@hosei.ac.jp,
    \\ \{jielei, ysli\}@mail.xidian.edu.cn, yuwing@gmail.com
}

\usepackage{bibentry}

\begin{document}

\maketitle

\begin{abstract}
In most video platforms, such as Youtube, and TikTok, the played videos usually have undergone multiple video encodings such as hardware encoding by recording devices, software encoding by video editing apps, and single/multiple video transcoding by video application servers. Previous works in compressed video restoration typically assume the compression artifacts are caused by one-time encoding. Thus, the derived solution usually does not work very well in practice. In this paper, we propose a new method, temporal spatial auxiliary network (TSAN), for transcoded video restoration. Our method considers the unique traits between video encoding and transcoding, and we consider the initial shallow encoded videos as the intermediate labels to assist the network to conduct self-supervised attention training. In addition, we  employ adjacent multi-frame information and propose the temporal deformable alignment and pyramidal spatial fusion for transcoded video restoration. The experimental results demonstrate that the performance of the proposed method is superior to that of the previous techniques. The code is available at \textcolor[RGB]{233,3,130}{\url{https://github.com/icecherylXuli/TSAN}}.

\end{abstract}

\begin{figure}[!h]
\centering
\includegraphics[width=0.45\textwidth]{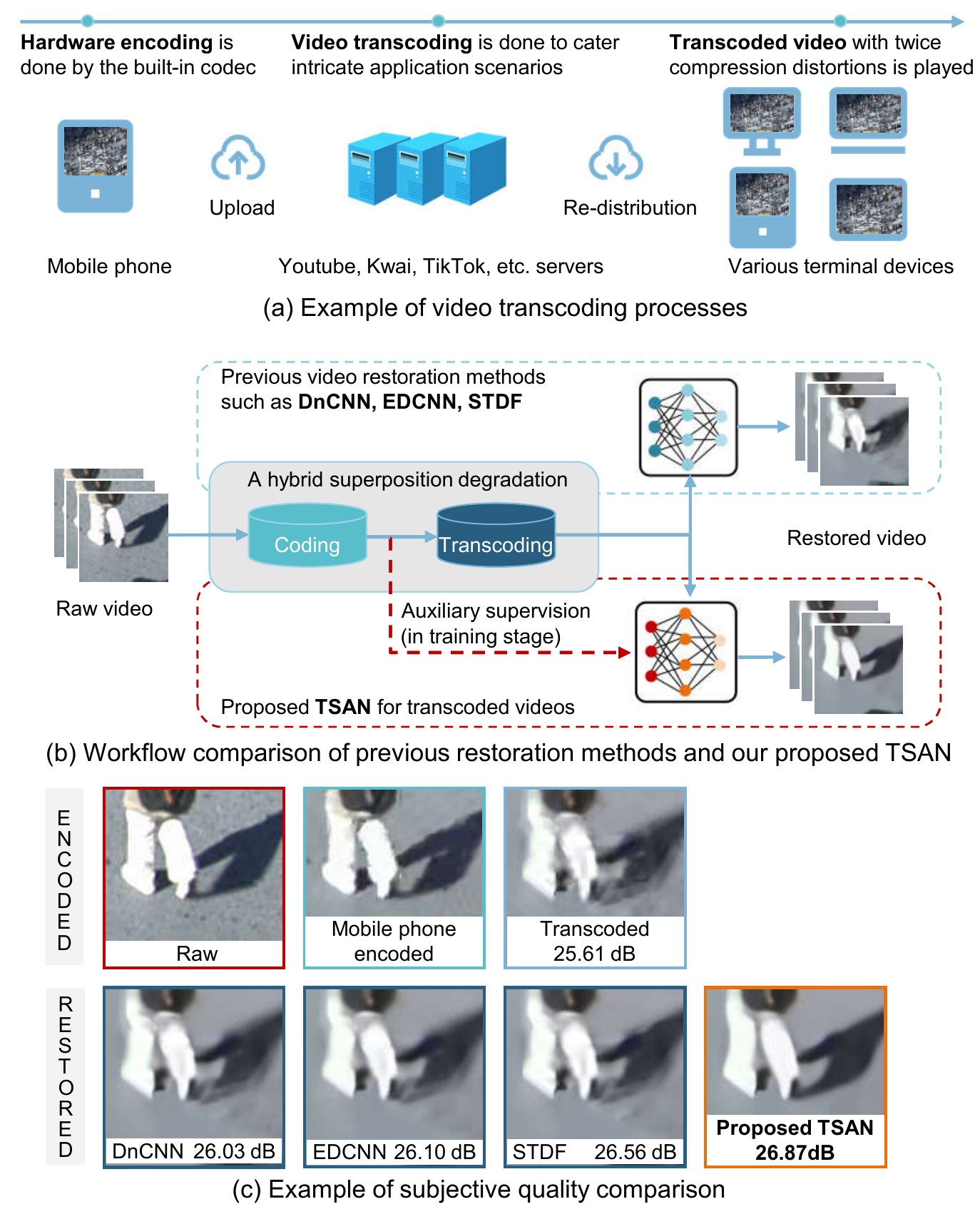}
\caption{(a) Example of video transcoding processes. (b) Workflow comparison of previous restoration methods and our proposed TSAN.  (c) Example of subjective quality comparison, including our TSAN, DnCNN \cite{DnCNN}, EDCNN \cite{EDCNN}, and STDF \cite{STDF}.}
\label{fig_transill}
\end{figure}

\section{Introduction}
Massive uncompressed video data is a substantial burden for hardware storage and transmission. Without proper encoding, it is almost impossible to transmit a high resolution video through the widespread 4G/5G network in real-time. Over the past decades, many video coding algorithms have been emerged to exploit video contents' spatial and temporal redundancy while pursuing an acceptable visual quality. The classic video coding standards, such as H.262/MPEG-2 \cite{rec1994h}, H.263 \cite{recommendation1998263}, H.264/AVC \cite{telecom2003advanced}, H.265/HEVC \cite{sullivan2012overview}, have been developed rapidly. Indeed, due to the limitations of transmission conditions and various mobile devices, almost all the videos we encounter on the Internet were transcoded to meet the target bitrates for efficient transmission. For example, a video taken directly using our mobile phone needs to be compressed at least twice (hardware encoding in mobile phone before uploading, and the transcoding in video server before re-distribution) before it can be shared on Youtube, Kuai or TikTok, etc., as illustrated in Fig. \ref{fig_transill} (a). Intuitively, transcoded videos that end users viewed usually suffer from single/multiple transcoding degradation. 

\par  
Recently, some learning-based methods have been explored to eliminate the artifacts of compressed images and videos. 
However, they focused on improving the quality of encoded videos without transcoding. The integral distortion by video encoding and transcoding is not a simple superposition of two coding distortions, as transcoding will further introduce different degrees of distortions in the degraded regions with severe or slight artifacts. In Fig.\ref{fig_transill} (c), only the road texture is lost in the mobile phone encoded image, but severe artifacts appear on the character's lower body and its shadow in the transcoded image. This makes the transcoded video restoration task difficult. To the best of our knowledge, there is currently no learning-based works dedicated to improving the quality of transcoded videos. A straightforward solution is to retrain these previous algorithms and reuse them to transcoded videos. Fig. \ref{fig_transill} (b) shows the workflow comparison of previous methods and our TSAN for transcoded videos. As demonstrated in Fig.\ref{fig_transill} (c), the performance of previous methods is limited because they are specifically designed for the video whose contents only suffer from single compression. Considering that the distortions of transcoded videos are a hybrid superposition of twice or more encoding distortions, we propose a temporal spatial auxiliary network to enhance transcoded videos. 

The main contributions can be summarized as follows: 
\begin{itemize}
\item This study is the first exploration on transcoded video restoration with deep neural networks. We reveal most videos suffer from transcoded deterioration and verify these previous learning-based restoration algorithms for video encoding are fragile for video transcoding.
\item We propose a network paradigm that uses the initial encoding information as an auxiliary supervised label to assist the network training. More specifically, we design auxiliary supervised attention and global supervised reconstruction modules to improve the algorithm performance in a coarse-to-fine manner. 
\item We design a temporal deformable module capable of offsetting the motion in a progressive motion-learning manner while extracting abundant valuable features. Furthermore, a pyramidal spatial fusion adopting four diverse downsampling filters is developed to capture more lossy details at multiple spatial scaling levels.
\item We quantitatively and qualitatively demonstrate our proposed method is superior to that of the previous methods.
\end{itemize}

\section{Related Work}  \label{sec2} 
\subsection{Encoded Image/Video Restoration}
Inspired by the success of deep learning, a large number of recent works \cite{liu2020deep, ARCNN, DnCNN, VRCNN, QECNN, he2018partition, ding2019switchable,TOF, MFQE, MFQE2.0, STDF} have shown that convolutional neural network (CNN) achieves excellent performance in enhancing quality of compressed images and videos. ARCNN \cite{ARCNN} is the first work to leverage CNN to reduce artifacts caused by JPEG. Next, benefitting from its robustness, DnCNN \cite{DnCNN} is often used as an image restoration baseline, including denoising, artifacts reduction, and so forth. As a learning-based post-processing method, VRCNN \cite{VRCNN} was developed to promote the HEVC intra coding frames' quality. Later, quality enhancement convolutional network (QECNN) was proposed to improve the quality for I frames and P/B frames of HEVC separately. Analyzing the characteristics of video coding, a partition-masked CNN \cite{he2018partition} was designed, which employed the decoded frames' partition information to instruct the network to improve performance. Meanwhile, \citet{EDCNN} designed an enhanced deep convolutional neural network (EDCNN) as an efficient in-loop filter to remove annoying artifacts and achieve a better quality of experience. 

Due to the lack of effective use of adjacent information which is capable of providing supplementary details, these single-image encoded restoration works can improve the quality of damaged frames, but their improvement abilities are limited. Multi-image encoded restoration algorithms are becoming a prevalent trend. In task-oriented flow (TOFlow) \cite{TOF}, the learnable motion estimation component is self-supervised to facilitate video restoration. Later, MFQE \cite{MFQE} and its extended version \cite{MFQE2.0} developed a PQF detector to search for the highest quality reference frame, thus improving the damaged videos' quality. Moreover, to handle motion relationships efficiently, a spatio-temporal deformable fusion scheme \cite{STDF} is proposed to aggregate temporal information so as to eliminate undesirable distortions.

\begin{figure*}[ht]
\centering
\includegraphics[width=\textwidth]{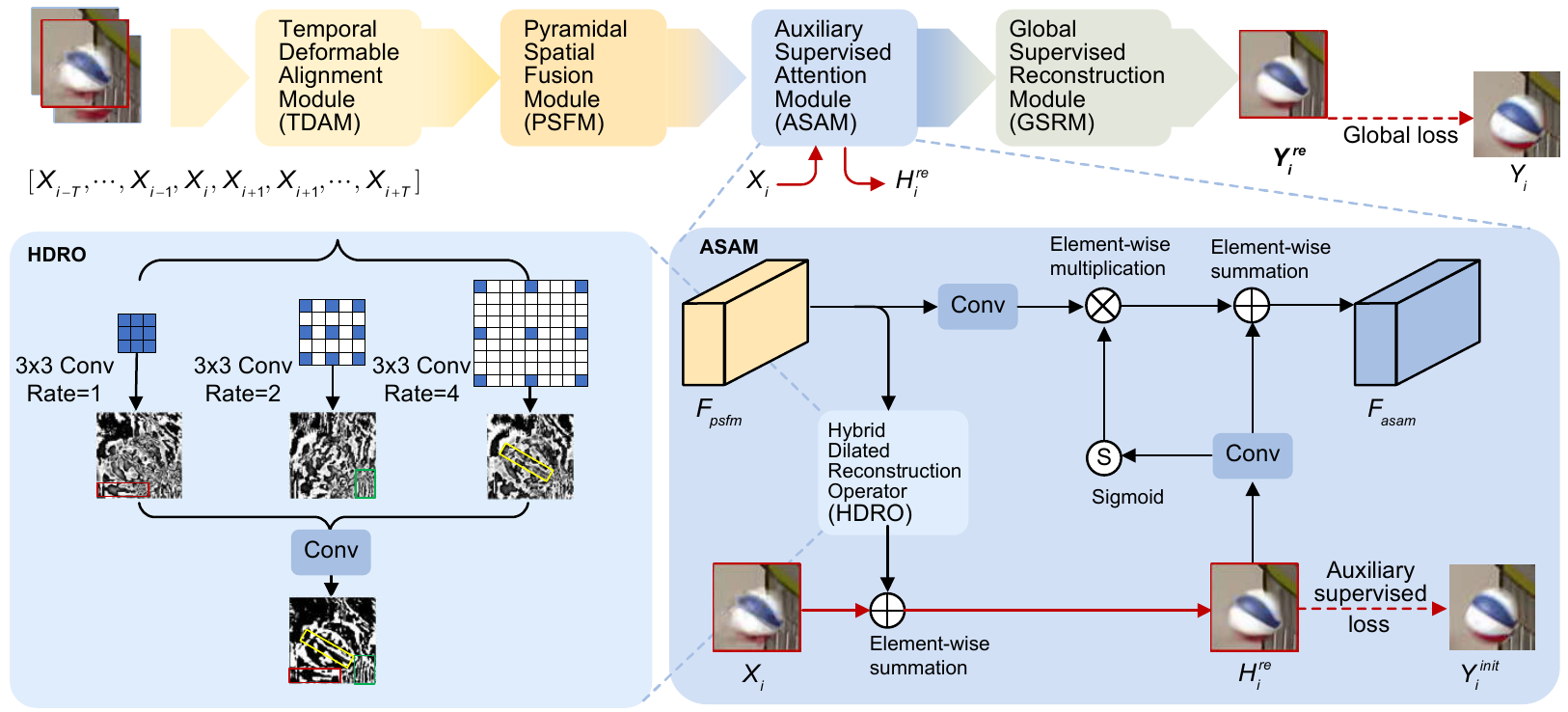}
\caption{Architecture of our temporal spatial auxiliary network (TSAN) and structure of auxiliary supervised attention module.  
} 
\label{fig-archi}
\end{figure*}

\subsection{Dilated Convolution}
Dilated convolutions \cite{yu2015multi}, also called atrous convolutions, can expand receptive fields while keeping the same resolution of feature maps. It is widely used in semantic segmentation \cite{chen2017deeplab, ASPP}, image classification \cite{yu2017dilated}, image restoration \cite{yu2018generative, guo2021efficientderain}, etc.  Atrous spatial pyramid pooling\cite{chen2017deeplab, ASPP} employed atrous convolution in cascaded or parallel by adopting multiple atrous rates and was used to handle the problem of semantic segmentation at different spatial scales. In image classification, dilated residual network \cite{yu2017dilated} was developed to improve the accuracy of downstream applications, and it outperforms its non-dilated counterparts. In \cite{yu2018generative}, a generative inpainting method utilized dilated convolution to rough out the missing contents.  \citet{guo2021efficientderain} proposed EfficientDeRain with a pixel-wise dilation filtering to predict multi-scale kernels for each pixel.

\subsection{Deformable Convolution}
Standard convolution is innately constrained in establishing geometric transformations on account of the invariable local receptive fields. The variant, deformable convolution (DConv) \cite{Dconv}, was developed to obtain learnable spatial information with the guidance of the additional offset. Temporally deformable alignment network \cite{TDAN} first applied DConv to align the neighboring frames instead of optical flow to predict high-resolution videos. Inspired by \cite{TDAN}, EDVR \cite{wang2019edvr} elaborated on a pyramid manner to estimate offset more precisely. Interlayer restoration network \cite{he2021interlayer} for scalable high efficiency video coding employed DConv to compensate the compression degradation difference through the multi-scale learnable offsets. In particular, \citet{chan2021understanding} revealed the relation between deformable alignment and flow-based alignment and proposed an offset-fidelity loss to alleviate the training instability.

\section{Methodology }  \label{sec3}
\subsection{Preliminary}
Transcoded videos usually suffer from multiple encoding degradations and are hard to be restored due to the complicated unknown process. Table \ref{tab_compare} summarizes the performance of the classic restoration method \cite{STDF} on one-time video encoding and transcoding in three test sequences. Note that the DNN-based method was retrained on the dataset  which covers the same videos but suffers from different degradation. The typical STDF was originally designed for improving the quality of the distorted videos compressed by the H.265/HEVC reference software HM16.5. Here we also follow the same configuration to prepare the training dataset. The procedure of preparing transcoding dataset will be depicted in Sec. 4.1. In Table \ref{tab_compare}, although the qualities (peak signal-to-noise, PSNR) of the three listed sequences suffering one-time encoding and transcoding fluctuate, they are similar. However, the improvement of video transcoding by STDF decreased from 1.051 dB to 0.646 dB in sequence \textit{BQMall}. Likewise, when applying STDF to video transcoding, different degrees of degradations are observed in all the listed sequences. These results validate that \textit{the previous learning-based methods for video encoding are fragile for video transcoding}.

\begin{table}[t]
  \centering
\resizebox{0.47\textwidth}{!}{  
    \begin{tabular}{ccccc}
    \hline
    \multirow{2}{*}{Sequences}              &\multirow{2}{*}{Status}      &One-time Encoding               &Transcoding          &\multirow{2}{*}{$\Delta$}       \cr
    							  					          &&(QP 37)           &(1 or 0.5 Mbps)  &\cr
      \hline
      \multirow{3}{*}{\makecell[c]{Basketball-\\Drill}}      &Before              &31.591              &31.747                    &-\cr 

                                                            &After                 &32.409              &32.303                    &-       \cr
                                                            \cline{2-5}
                                                            &$\Delta$PSNR &0.818                &0.556                      &\bf{-0.262}       \cr
      \hline
      \multirow{3}{*}{BQMall}                 &Before              &31.297              &31.949                    &-\cr 

                                                            &After                 &32.340              &32.595                   &-       \cr
                                                            \cline{2-5}
                                                            &$\Delta$PSNR &1.051                &0.646                      &\bf{-0.405}       \cr
      \hline
      \multirow{3}{*}{BQTerrace}          &Before              &31.247                &31.565                    &-\cr 

                                                            &After                 &31.894              &31.954                    &-       \cr
                                                            \cline{2-5}
                                                            &$\Delta$PSNR &0.647                &0.389                      &\bf{-0.258}       \cr
    \hline
    \end{tabular}}
        \caption{Comparison of restoration on one-time encoding and transcoding in terms of PSNR (dB).  ``Before'' status denotes the sequence has been compressed but not enhanced. ``After'' denotes the sequence has been enhanced by the retrained STDF \cite{STDF}.} 
    \label{tab_compare}
\end{table}

\subsection{Network Architecture Overall}
    As discussed above, the artifacts of compressed videos are mostly a combination of video encoding and transcoding distortions. Hence, we design a progressive restoration network that divides the transcoded video restoration task into two parts and restores the transcoded videos in a coarse-to-fine manner. The first part focuses on removing the distortion introducing by transcoding at a lower bitrate. Then, the second part is inclined to eliminate the artifacts caused by the initial video encoding. For this purpose, \textit{we employ the initial encoded video with a high bitrate as an additional auxiliary supervised label to assist network at training stage.} We denote our method, temporal spatial auxiliary network (TSAN). 

Given $2T+1$ consecutive low-quality frames $X_{[i-T:t+T]}$, we denote the center frame $X_i$ as the target frame need to be restored and the other frames as the reference frames. The input of the network is the target frame $X_i$ and the $2T$ neighboring frames, and the output is the enhanced target frame $Y_i^{re}$. The objective function can be formulated as follows,
\begin{equation}  \label{equ_overall}
    Y^{re}_i =  {\bf Net_{TSAN}}(X) ,  
\end{equation}
\noindent
where ${\bf {Net_{TSAN}}}$ is our proposed temporal spatial auxiliary network and $X$ is a stack of transcoded frames which is defined as
\begin{equation}  \label{equ_overall}
    X =  [X_{i-T},\cdots,X_{i-1},X_i, X_{i+1},\cdots,X_{t+T}],  
\end{equation}
where $i$ denotes the frame number and $T$ is the maximum number of reference frames. 

 The architecture of TSAN is shown in Fig. \ref{fig-archi}. TSAN is devised to estimate a high quality output with the guidance of its consecutive transcoded frames. In the following subsection, we will give detailed analysis on the motivation and rationality of each module.

\subsection{Auxiliary Supervised Attention}
The temporal deformable alignment and pyramidal spatial fusion modules serve for the severe distortions where the contents have been degraded repeatedly. However, the lossy information is hard to be recovered due the hybrid transcoding degradation. Combining the traits of video encoding and transcoding, we proposed an auxiliary supervised attention module (ASAM) whose structure is illustrated in the blue part of Fig. \ref{fig-archi}. First, we use a hybrid dilation reconstruction operator (HDRO) to predict the high-frequency map. Specifically, we apply three dilated convolutions with different dilation rates ($r = 1, 2, 4$) to reconstruct the lossy frequency maps at different spatial scale. Intuitively, the $3\times3$, $5\times5$, and $7\times7$ receptive fields of each pixel are supported by these convolutions and the various highlighted texture, like the red, green, and yellow rectangular boxes, are generated. Note that the 1-dilated convolution is equivalent to the standard convolution. Following the parallel dilated convolutions, these frequency maps are sent into a $3\times3$ standard convolution to yield the integrated result. The process of HDRO is given by:
\begin{equation}  \label{equ_overall}
F_{HDRO} = Rec([D_{1}(F_{psfm}),D_{2}(F_{psfm}),D_{4}(F_{psfm})]),
\end{equation}
where $D_{1}$, $D_{2}$, and $D_{4}$ denote dilated convolutions with 1, 2 and 4 dilation rates, respectively. $Rec(\cdot)$ is the $3\times3$ standard convolution.

After HDRO, the integrated frequency map is sum up to the low-quality target frame $X_i$ to yield the initial restored one $H_i^{re}$.
Up to now, the initial restoration stage is complete. Next, we feed $H_i^{re}$ again into the convolution for providing excited features. The sigmoid activation function is used to restrict these features in [0, 1] and generate supervised attention maps. Then, these earliest input feature through a $3\times3$ transitional convolution and are refined by the supervised attention maps. Ultimately, these self-refined features are added to the excited features by $H_i^{re}$ as the output of module. 

In summary, ASAM plays an essential role in guiding the first part of our network to approach the intermediate lossy representation and establishing a connection between the transcoding degradation and initial encoding degradation. In the second part, we design a global supervised reconstruction module (GSRM) which consists of 10 residual blocks and the HDRO with a short-cut connection of target frame $X_i$. With the help of GSRM, the final restored output $Y_i^{re}$ is reconstructed.

\begin{figure}[th]
\centering
\includegraphics[width=0.5\textwidth]{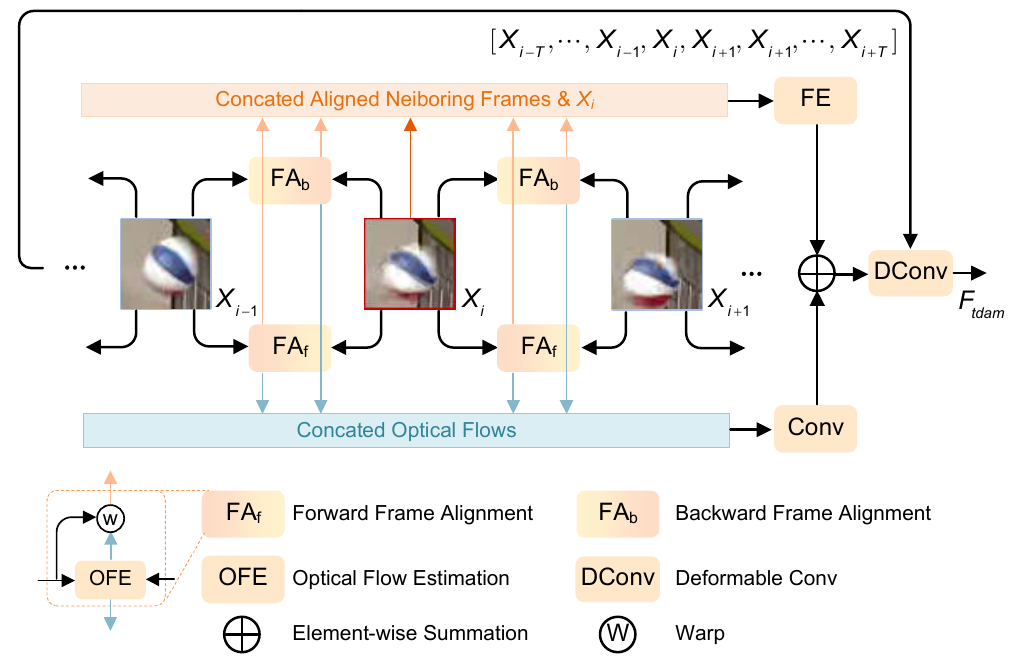}
\caption{Structure of temporal deformable alignment.} 
\label{fig_tdam}
\end{figure}

\subsection{Temporal Deformable Alignment}
Adjacent frames are essential for target frame restoration but they are not equally informative due to view angle, motion blocking, and video compression problems. Hence, we leverage the neighboring forward and backward frames to exploit more advantageous information. In Fig. \ref{fig_tdam}, we employ optical flow estimation (OFE), i.e. SPyNet \cite{spynet}, to compute the forward and backward optical flows among the adjacent frames. Then, these estimated optical flows are regarded as the plain motion information and are warped with the input frames to yield the plain aligned frames. Here we describe the forward frame alignment in detail, and the backward frame alignment can be inferred similarly. The forward alignment at $i\!-\!t\!-\!1$ timestep is done, 
\begin{equation}  \label{equ_overall}
    X^A_{i\textnormal{-}t\textnormal{-}1\to i\textnormal{-}t} = warp({\rm \bf OFE}(X_{i-t-1},X_{i-t}), X_{i-t-1}),
\end{equation}
where $t\!\in\![0,T)$ and ${\rm \bf OFE}(\cdot)$ is optical flow estimation. We use bilinear interpolation to implement $warp(\cdot)$ function. 

Later, these initial aligned frames $X^A$ and the target frame $X_i$ are concatenated together to send into feature excitation (FE) and generate the motion refinements on the basis of plain motion estimation. Three stacked plain $3\times3$ convolution layers are adopted to feature excitation. Integrating the plain motion transformed by a convolutional filter and the motion refinements,  the more progressive refined motion information is generated, and it is regarded as the learnt predicted offset $\bigtriangleup P$ to help the explicit temporal deformable alignment. The mathematical equation is
\begin{equation}
    F_{tdam}(p_0) = \sum\limits_{p_k \in \mathcal{R} }{\omega_k \cdot X(p_0+p_k+\bigtriangleup p_k)},
\end{equation}
the deformable convolution will be operated on the deformed sampling locations $p_k\!\!+\!\! \bigtriangleup p_k$, where $\omega_k$ and  $p_k$  denote the weight and predicted offset for $k\textnormal{-}th$ location in  \begin{math}\mathcal{R}=\{(-1,-1),(-1,0),\cdots, (0,0),\cdots,(0,1),(1,1)\}\end{math}. Note that the $\bigtriangleup p_k\!\in\! \bigtriangleup P$. Finally, a stack of temporal deformable alignment features $F_{tdam}$ are acquired by compensating the motion in a progressive motion-learning manner.

\begin{figure}[t]
\centering
\includegraphics[width=0.47\textwidth]{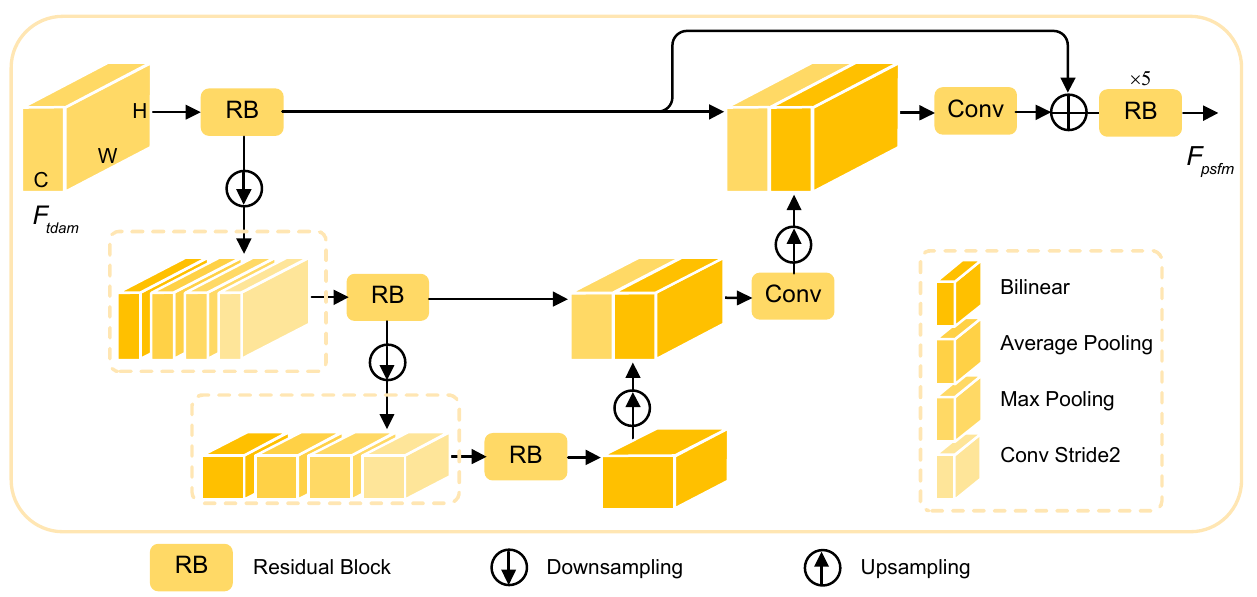}
\caption{Structure of pyramidal spatial fusion.
} 
\label{fig-psfm}
\end{figure}

\subsection{Pyramidal Spatial Fusion}
After obtaining the temporal deformable aligned features $F_{tdam}$, we design a pyramidal spatial fusion module (PSFM) based on an UNet \cite{UNet} structure to learn the contextual information at multiple spatial levels. The notable differences between our PSFM and UNet are two-fold. 

First, four diverse downsampling filters, including the bilinear operator, average pooling, max pooling, and strided convolutional filters, are adopted instead of single downsampling method. Following this pyramidal downsampling manner, we enlarge the aligned features' receptive filed to merge the neighboring information from the temporal dimension to spatial dimension and capture more lossy details at multiple scaling levels. Second, residual learning with a single shortcut connection is introduced to replace the plain $3\times 3$ convolutions. This variant is conducive to propagate the proceeding learnt valuable information from the shallower layers to the deeper ones, and ameliorate gradient vanishing and explosion. After that, we employ 5 residual blocks to generate enhanced features.

\subsection{Loss Function} \label{lossfunction}
We develop a loss function which consists of auxiliary supervised loss function and global supervised loss function. The initial shallow encoded video $Y_i^{init}$ with a high bitrate is regarded as the auxiliary supervised label, and the partial loss function is calculated as:
\begin{equation} 
    Loss_a = 1/N\sum\limits_{i=1}^N{\left \| Y_i^{init} - H_i^{re}\right \|^2},
\end{equation}
where $N$ is the batch size and MSE loss is adopted for optimization. Meanwhile, the global supervised loss function is calculated as:
\begin{equation}
    Loss_g = 1/N\sum\limits_{i=1}^N{\left \| Y_i - Y_i^{re}\right \|^2},
\end{equation}
where $Y_i$ denotes the raw target frame without any compression and $Y_i^{re}$ denotes the enhanced one. 

Our network is an end-to-end method and the two loss functions are combined together as the final loss of the entire network for back propagation process. It is defined as:
\begin{equation}
    Loss = \alpha \cdot Loss_a + \beta \cdot Loss_g,
\end{equation}
where $\alpha$ and $\beta$ are the weight factors. Validating by the related experiment, we set $\alpha = 0.2$ and $\beta = 0.8$.

\begin{table}[t]
\small
  \centering
    \setlength{\tabcolsep}{1.5mm}{   
    \begin{tabular}{cccc}
    \hline
    \multicolumn{2}{c}{Settings}&Initial Encoding&One-time Transcoding\cr

    \hline
     \multirow{2}{*}{Bitrate}&HR$^1$ &10 Mbps &1000 kbps\cr
     \cline{2-4}
    &LR$^2$ &10 Mbps &500 kbps\cr
        \hline
    \multicolumn{2}{c}{Coding Software}&X265 &X265\cr

    \multicolumn{2}{c}{Preset}  &Medium &Medium\cr

    \multicolumn{2}{c}{Rate Control Mode}  &Average Bitrate &Average Bitrate\cr

    \multicolumn{2}{c}{Loop Filters}  &Deblock and SAO&Deblock and SAO \cr
    \multicolumn{2}{c}{Group of Pictures}  &250 &250 \cr
    \multicolumn{2}{c}{Max References}  &3&3 \cr
    \hline
    \end{tabular}}
       \makecell[l]{\footnotesize{$^1$HR denotes high resolution videos higher than 720p. }\\
       \footnotesize{$^2$LR denotes low resolution videos lower than or equal to 720p. }\\}
\caption{Detailed settings of video initial encoding and one-time transcoding. The bitrate is set according to the resolution of videos and the other settings are same.}
\label{tab_dataset}
\vspace{-0.2cm}
\end{table}

\section{Experiments and Analyses }  \label{sec4}
\subsection{Experimental Settings} \label{sec41}
\textbf{Training and Testing Dataset.} To establish a training dataset for video transcoding restoration, we employed 108 sequences from Xiph.org \cite{Xiph}, VQEG \cite{VQEG}, and Joint Collaborative Team on Video Coding (JCT-VC) \cite{bossen2013common}. The resolutions of these sequences cover SIF, CIF, 4CIF, 360p, 480p,1080p, and 2k. We adopt all 18 standard test sequences from JCT-VC for testing.

\begin{table*}[ht]
  \centering
  \small
\resizebox{\textwidth}{!}{  
    \begin{tabular}{cccccccc}
    \hline	                                                   
    \multirow{2}{*}{Classes}&\multirow{2}{*}{Sequences}&DnCNN&QECNN&EDCNN&STDF&Proposed\cr
    &&(Zhang et al., TIP' 17)                    &(Yang et al., TCSVT' 18)               &(Pan et al., TIP' 20)                 & (Deng et al., AAAI' 20)  &  TSAN\cr
   \hline
    \multirow{2}{*}{A}
    &Traffic                       &0.309/0.005&0.238/0.004&0.292/0.005&0.392/0.009           &0.554/0.008\cr
    &PeopleOnStreet       &0.423/0.027&0.302/0.019&0.490/0.030&0.950/0.046            &1.260/0.054\cr
    \hline
     \multirow{5}{*}{B}
    &BasketballDrive       &0.375/0.012&0.275/0.010&0.429/0.013&0.600/0.016            &0.863/0.020\cr
    &BQTerrace               &0.299/0.007&0.280/0.005&0.332/0.007&0.389/0.008            &0.545/0.010\cr
    &Cactus                     &0.320/0.010&0.240/0.008&0.342/0.010&0.480/0.012            &0.743/0.017\cr
    &Kimono                    &0.263/0.010&0.196/0.008&0.280/0.011&0.368/0.012            &0.632/0.017\cr
    &ParkScene              &0.171/0.008&0.132/0.006&0.177/0.007&0.254/0.010            &0.447/0.015\cr
    \hline
    \multirow{4}{*}{C}
    &BasketballDrill         &0.448/0.011&0.339/0.008&0.484/0.011&0.556/0.013             &0.796/0.017\cr
    &BQMall                   &0.462/0.011&0.360/0.009&0.468/0.011&0.646/0.014             &0.985/0.019\cr
    &PartyScene            &0.241/0.012&0.174/0.009&0.264/0.012&0.360/0.016            &0.539/0.022\cr
    &RaceHorses           &0.274/0.014&0.189/0.011&0.299/0.015&0.417/0.018            &0.707/0.026\cr
   \hline
   \multirow{4}{*}{D}
   &BQSquare              &0.443/0.004&0.357/0.002&0.564/0.004&0.613/0.004                 &1.264/0.011\cr
   &BasketballPass      &0.366/0.007&0.373/0.004&0.457/0.005&0.675/0.009	          &0.948/0.013\cr
   &BlowingBubbles     &0.329/0.007&0.253/0.005&0.338/0.007&0.506/0.010		  &0.804/0.014\cr
   &RacesHorses         &0.474/0.007&0.385/0.005&0.442/0.006&0.501/0.008		  &0.826/0.013\cr
                                
    \hline
    \multirow{3}{*}{E}
       &FourPeople       &0.484/0.004&0.408/0.003&0.390/0.004&0.708/0.005              &0.933/0.006\cr
       &Johnny                &0.256/0.003&0.129/0.002&0.099/0.002&0.291/0.003              &0.480/0.004\cr
       &KristenAndSara   &0.358/0.003&0.293/0.003&0.155/0.003&0.521/0.004             &0.756/0.005\cr 
    \hline
    \multicolumn{2}{c}{\bf Average}          &0.350/0.009&0.274/0.007&0.350/0.009&0.513/0.012             &0.782/0.016\cr
   \hline
    \end{tabular}}
\caption{  Improvement (\textbf{$\Delta$PSNR/$\Delta$SSIM}) of our TSAN and previous DNN-based methods in  video transcoding.}
\label{tab_psnr}
\end{table*}

\textbf{Encoding and Transcoding Settings.} All videos in the training and testing dataset have been processed by initial encoding and one-time transcoding. The detailed settings have been listed in Table \ref{tab_dataset}. The video resolutions have been kept the same and the same encoding tool x265 \cite{x265} has been adopted in both the initial encoding and one-time transcoding. Both x265 presets have been set as ``medium'' and others settings including rate control strategy, group of pictures (GOP) size, loop filters, etc. are all the same default ones except the bitrates for them. The initial encoding bitrate is set as the high bitrate 10 Mbps. This is because we calculated that the average bitrate of 100 videos taken by iPhone12 is close to 10 Mbps. Meanwhile, the transcoding bitrate is set as 500/1000 kbps according to different resolutions to simulate the cases of real practical applications such as TikTok. We have randomly downloaded about 200 hundred videos of TikTok and do the statistical bitrate data for them.

\par
\textbf{Implementation Details.} We implement our TSAN with Pytorch 1.6.0 framework on a NVIDIA GeForce 2080Ti GPU and update it with Adam optimizer. The batch size is set to 16 and the learning rate is initialized as 1e-4. The network training stops after 300k iterations.

\subsection{Transcoding Restoration Performance}  \label{sec4-B}
\par 
Table \ref{tab_psnr} shows the performance of our method in transcoded videos, compared with DNN-based methods \cite{DnCNN, QECNN, EDCNN, STDF}. The delta peak signal-to-noise ratio ($\bigtriangleup$PSNR) and delta structural similarity index metric ($\bigtriangleup $SSIM) are calculated.

 \begin{figure*}[htb]
\centering
\includegraphics[scale=1]{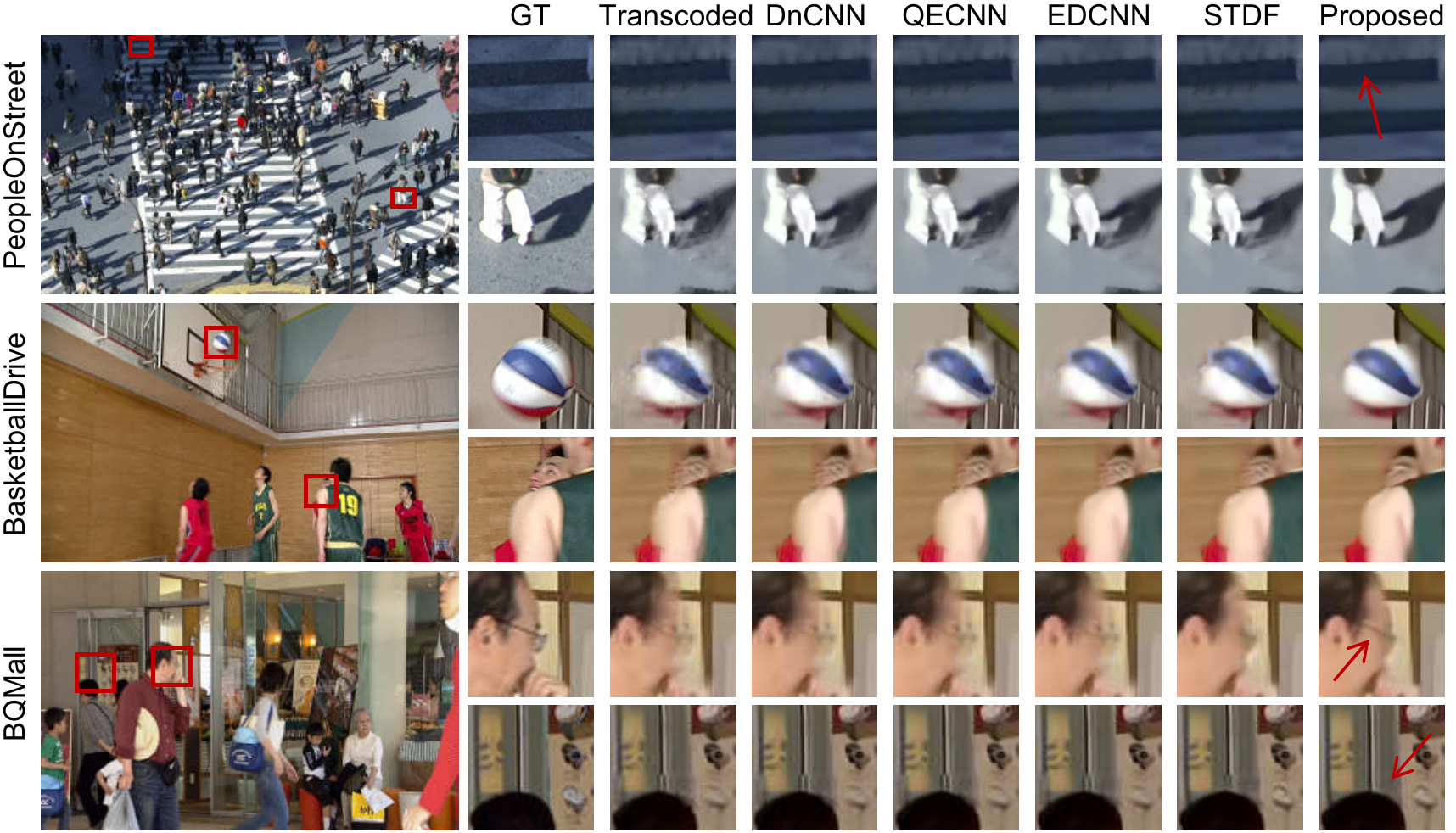}
\caption{Subjective quality performance of our method and previous methods. }
\label{fig_sub}
\end{figure*}

\begin{table}[!]
  \centering
\resizebox{0.48\textwidth}{!}{  
    \begin{tabular}{cccc}
    \hline
    Components&V1&V2&Proposed\cr
   \hline
    Temporal Deformable Alignment &\checkmark&\checkmark&\checkmark \cr
    Pyramidal Spatial Fusion&$\times$&\checkmark  &\checkmark  \cr
    Auxiliary Supervised Attention&$\times$&$\times$ &\checkmark \cr
        \hline
    \end{tabular}}
       \caption{Compositions of different proposed networks.}
 \label{tab_component}
\end{table}
\begin{table}[h]
  \centering
  \small
\resizebox{0.47\textwidth}{!}{  
    \begin{tabular}{cccc}
    \hline
    Sequences&V1&V2&Proposed\cr
    \hline
    Traffic                       &0.434/0.012&0.462/0.007       &0.554/0.008\cr
     PeopleOnStreet       &0.913/0.042&1.084/0.048      &1.260/0.054\cr
    \hline
    BasketballDrive       &0.633/0.016&0.687/0.017       &0.863/0.020\cr
    BQTerrace               &0.415/0.008&0.474/0.009       &0.545/0.010\cr
    Cactus                     &0.533/0.012&0.605/0.014       &0.743/0.017\cr
    Kimono                    &0.468/0.014&0.528/0.015       &0.632/0.017\cr
    ParkScene              &0.329/0.011&0.356/0.013       &0.447/0.015\cr
    \hline
      BasketballDrill       &0.566/0.013&0.607/0.014       &0.796/0.017\cr
      BQMall                  &0.689/0.014&0.783/0.016       &0.985/0.019\cr
       PartyScene          &0.353/0.016&0.401/0.018       &0.539/0.022\cr
       RaceHorses         &0.537/0.021&0.551/0.023       &0.707/0.026\cr
       \hline
      BQSquare              &0.732/0.005&1.023/0.009      &1.264/0.011\cr
      BasketballPass      &0.608/0.007&0.861/0.012      &0.948/0.013\cr
      BlowingBubbles     &0.531/0.010&0.634/0.012      &0.804/0.014\cr
      RacesHorses         &0.487/0.007&0.760/0.012      &0.826/0.013\cr
     \hline
     FourPeople             &0.703/0.005&0.825/0.006      &0.933/0.006\cr
     Johnny                    &0.352/0.003&0.520/0.004      &0.480/0.004\cr
     KristenAndSara      &0.589/0.004&0.747/0.005      &0.756/0.005\cr 
    \hline
    \bf Average              &0.548/0.012&0.662/0.014        &0.782/0.016\cr
    \hline
    \end{tabular}}
\caption{Ablation study in terms of improvement (\textbf{$\Delta$PSNR/$\Delta$SSIM}).}
\label{tab_ablation}
\end{table}

The average PSNR gain is 0.782 dB when the resolutions of videos vary from 416$\times$240 to 2560$\times$1600, and the highest PSNR gain is 1.264 dB in \textit{BQSquare}. It demonstrates that the proposed network can substantially improve the quality of transcoded videos. To further validate the effectiveness of our TSAN, we reimplement the four representative video restoration DNN-based methods \cite{DnCNN, QECNN,EDCNN,STDF}. Note that the presented results are generated by the networks trained with the transcoded dataset. As can be observed, with the help of these methods, the averages of $\bigtriangleup$PSNR are 0.274$\sim$0.513 dB on the 18 test videos suffered from video encoding and transcoding. It should be noted that these previous methods have an essential effect on the video restoration of single encoding. However, when applied to transcoded video restoration, their utilities are decreased. Comparing with these previous methods for one-time compression scenarios, our TSAN can improve the average $\bigtriangleup$PSNR by 0.269$\sim$0.508 dB. We deem that this is due to the different application scenarios and well-designed  network. In terms of SSIM, the average gain of proposed TSAN is 0.016, which is about 0.004$\sim$0.009 more than previous methods. The results verify that our proposal can generate a delightful perceptual quality improvement in comparison to previous DNN-based methods. The parameter of our method is 5.75M, while the parameter of the previous method is 0.56$\sim$3.84M. Despite achieving higher quality performance, the efficiency of our algorithm should be optimized.

\begin{table}[htb]
  \centering
    \begin{tabular}{cccc}
    \hline
    ($\alpha$, $\beta$)&(0, 1)&(0.2, 0.8)&(0.5, 0.5)\cr
   \hline
    $\Delta$PSNR (dB)&0.586&0.757&0.703 \cr
    \hline
    \end{tabular}
       \caption{Comparison of different weight factors in TSAN loss function, Class C.}
 \label{tab_loss}
\end{table}

\subsection{Ablation Study}  \label{sec4-C}
To validate the contribution of each component, a baseline combining components gradually is presented in Table \ref{tab_component}. We develop three variants of  TSAN: the first version (V1) only consists of a temporal deformable alignment module; the second version (V2) is extended by pyramidal spatial fusion module; the final version (Proposed) includes not only the proceeding parts and but also the auxiliary supervised attention and global supervised reconstruction modules. 

In practice, we listed the corresponding improvement in terms of $\Delta$PSNR and $\Delta$SSIM in Table \ref{tab_ablation}. As shown in it, the performance gains of the three versions increase gradually and steadily, and the highest performance is obtained by our final version which is well-designed for transcoding videos. Notably, the average PSNR gain of V1 is 0.548 dB, and it outperforms those of the previous DNN-based methods (0.274$\sim$0.513 dB). Compared with previous methods mentioned above, the baseline of our proposal is more capable of eliminating the artifacts under the circumstance that the neighboring information is leveraged fully and more precise motion alignment is performed. Since the pyramidal spatial fusion scheme further explores the lossy contextual details, it brings a 20.8\% increment compared with V1. It is apparent that the effectiveness of PSFM has been demonstrated. 
Furthermore, we verify the significant advantages of introducing auxiliary supervised attention. As mentioned above in Sec. \ref{lossfunction}, $\alpha$ and $\beta$ are the weight factors that can control the proportion of auxiliary supervised loss function and global supervised loss function in the whole one. When $\alpha$ = 0 and $\beta$ = 1, the network removed ASAM, i.e., V2. From Table \ref{tab_loss}, we can observe that utilizing the initial encoded video with high bitrate is conducive to learning more lossy information and reduce annoying distortions. More specifically, a 1:4  combination of the auxiliary supervised loss function and global supervised one can achieve a higher result. 
Note that the weight factors are not optimal, because we deem that further optimization of the weight factors will bring limited benefits, and we do not conduct too many experiments to optimize it.

\par
\subsection{Subjective Performance} 
Fig. \ref{fig_sub} shows the subjective quality performance of our TSAN and previous method \cite{DnCNN, QECNN, EDCNN, STDF} for transcoding restoration. Comparing the highlighted area of \textit{PeopleOnStreet}, we can find that the severely distorted zebra crossing is restored and the artifacts of the shadow are removed by our method. Likewise, the texture and graininess including the basketball and basketball player's face in sequence \textit{BasketballDrive} and the glasses frame and metal rod in sequence \textit{BQMall} are restored to a great extent. According to the favorable subjective quality performance, we can conclude that our method can acquire not only substantial objective achievements but also pleasuring perceptual results.

\section{Conclusion}  \label{sec5}
In this paper, we first explore the connection and difference between one-time video encoding and transcoding. Then, we demonstrate that these previous learning-based restoration methods are not robust for video transcoding. Based on this, we proposed a network paradigm that take advantage of initial encoding information as a forepart label to instruct the network optimization. Specifically, we proposed a temporal spatial auxiliary network (TSAN) which including temporal deformable alignment, pyramidal spatial fusion, and auxiliary supervised attention mainly to improve transcoded videos. This work is the first time dedicated to transcoded video restoration and we believe that this work can arouse broad interest in video restoration community.

\nobibliography{aaai22}

%
%
\bibliographystyle{aaai}
\bibliography{TSAN}

\end{document}